\documentclass[a4paper,fleqn]{cas-dc}

\usepackage[numbers]{natbib}
\usepackage{xcolor}
\usepackage{subcaption}

\begin{document}
\let\WriteBookmarks\relax
\def\floatpagepagefraction{1}
\def\textpagefraction{.001}

\shorttitle{Continual learning for rotating machinery fault diagnosis with cross-domain environmental and operational variations}

\shortauthors{Risca et~al.}

\title [mode = title]{Continual learning for rotating machinery fault diagnosis with cross-domain environmental and operational variations}


\tnotetext[1]{Work funded by Portuguese Foundation for Science and Technology under project doi.org/10.54499/UIDP/00760/2020 and Ph.D. scholarship PRT/BD/154713/2023. It also received EU funds, through Portuguese Republic’s Recovery and Resilience Plan, within project PRODUTECH R3.}

%

\author[1]{Diogo Risca}[orcid=0009-0000-1495-0662]

\author[1]{Afonso Louren\c{c}o}[orcid=0000-0002-3465-3419]

\cormark[1]


\ead{fonso@isep.ipp.pt}



\affiliation[1]{organization={GECAD, ISEP, Polytechnic of Porto},
    addressline={Rua Dr. António Bernardino de Almeida}, 
    city={Porto},
    postcode={4249-015}, 
    country={Portugal}}
   


\author[1]{Goreti Marreiros}[orcid=0000-0003-4417-8401]

\cortext[cor1]{Corresponding author}


\begin{abstract}
Although numerous machine learning models exist to detect issues like rolling bearing strain and deformation, typically caused by improper mounting, overloading, or poor lubrication, these models often struggle to isolate faults from the noise of real-world operational and environmental variability. Conditions such as variable loads, high temperatures, stress, and rotational speeds can mask early signs of failure, making reliable detection challenging. To address these limitations, this work proposes a continual deep learning approach capable of learning across domains that share underlying structure over time. This approach goes beyond traditional accuracy metrics by addressing four second-order challenges: catastrophic forgetting (where new learning overwrites past knowledge), lack of plasticity (where models fail to adapt to new data), forward transfer (using past knowledge to improve future learning), and backward transfer (refining past knowledge with insights from new domains). The method comprises a feature generator and domain-specific classifiers, allowing capacity to grow as new domains emerge with minimal interference, while an experience replay mechanism selectively revisits prior domains to mitigate forgetting. Moreover, nonlinear dependencies across domains are exploited by prioritizing replay from those with the highest prior errors, refining models based on most informative past experiences. Experiments show high average domain accuracy (up to 88.96\%), with forgetting measures as low as \(2.70 \times 10^{-3}\) across non-stationary class-incremental environments.
\end{abstract}


\maketitle

\section{Introduction}

In the era of rapid industrial development, the diagnosis of key components of machinery has become increasingly crucial. Rotating machinery, a central hub for power and energy transmission, is indispensable in engineering fields, including industrial, automotive, marine, and aerospace applications \citep{zhou2023rotating}. Bearings, critical components of rotating equipment, are expected to operate continuously under challenging conditions such as variable loads, high stress, elevated temperatures, and high rotational speeds \citep{wang2022multisource}. Fault conditions, which are generally caused by improper mounting, overloading, or improper lubrication, can significantly influence bearing strain and deformation. Thus, resulting in performance degradation, excessive vibration, noise, and secondary damage to other components if left unchecked, leading to approximately 50\% of total machine failures. Therefore, early detection of defects is a high priority for condition monitoring, as it allows scheduled maintenance before severe mechanical damage, catastrophic accidents, and operational downtime \citep{tama2023recent}.

\begin{figure}[ht]
    \centering
    \includegraphics[width=0.4\textwidth]{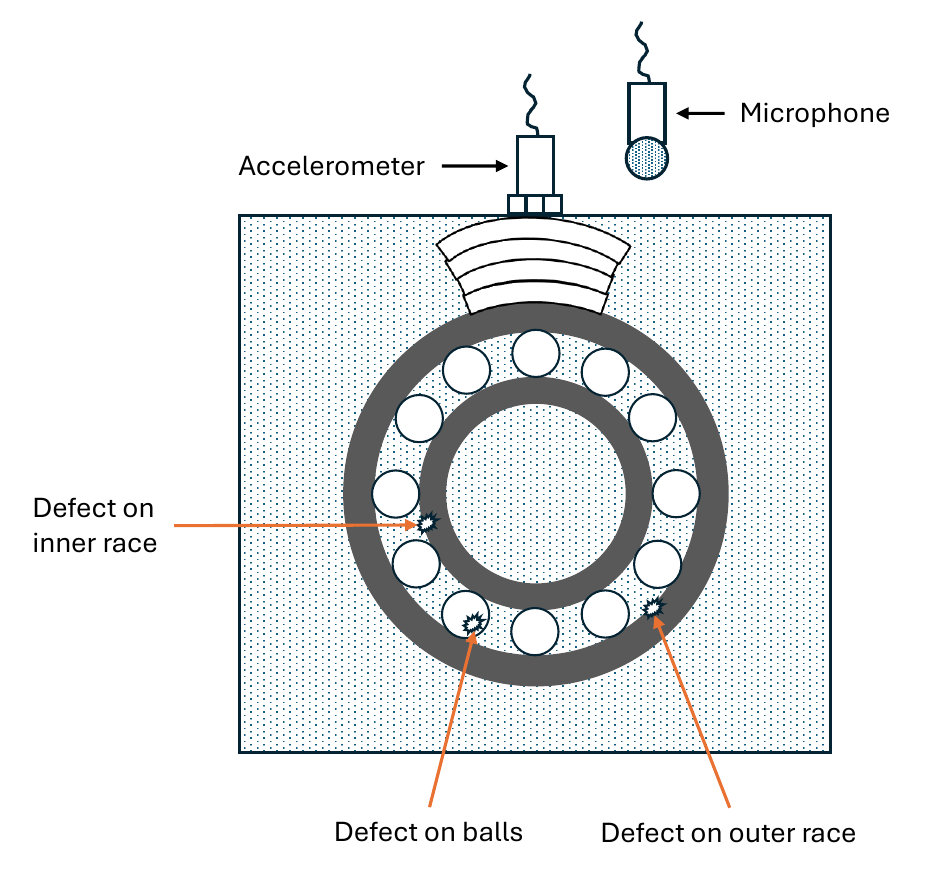}
    \caption{Accelerometer in radial direction on central axis of bearing housing, and microphone in near field condition}
    \label{fig:sensor}
\end{figure}

\textbf{Measurement techniques.} For this purpose, in recent decades, various techniques have been developed for monitoring and diagnosing rolling bearings, mainly based on vibration signals and acoustic emission, as illustrated in Figure \ref{fig:sensor}. First, exploiting transient elastic waves when deformation occurs within a material, from which the sudden release of strain energy could span a wide range of frequencies \citep{Balerston1969}, and then exploring the causes, influences, styles, and generating mechanisms of both these measurement techniques \citep{Hawman1988, mba2006development}.

\begin{figure}[ht]
    \centering
    \includegraphics[width=0.5\textwidth]{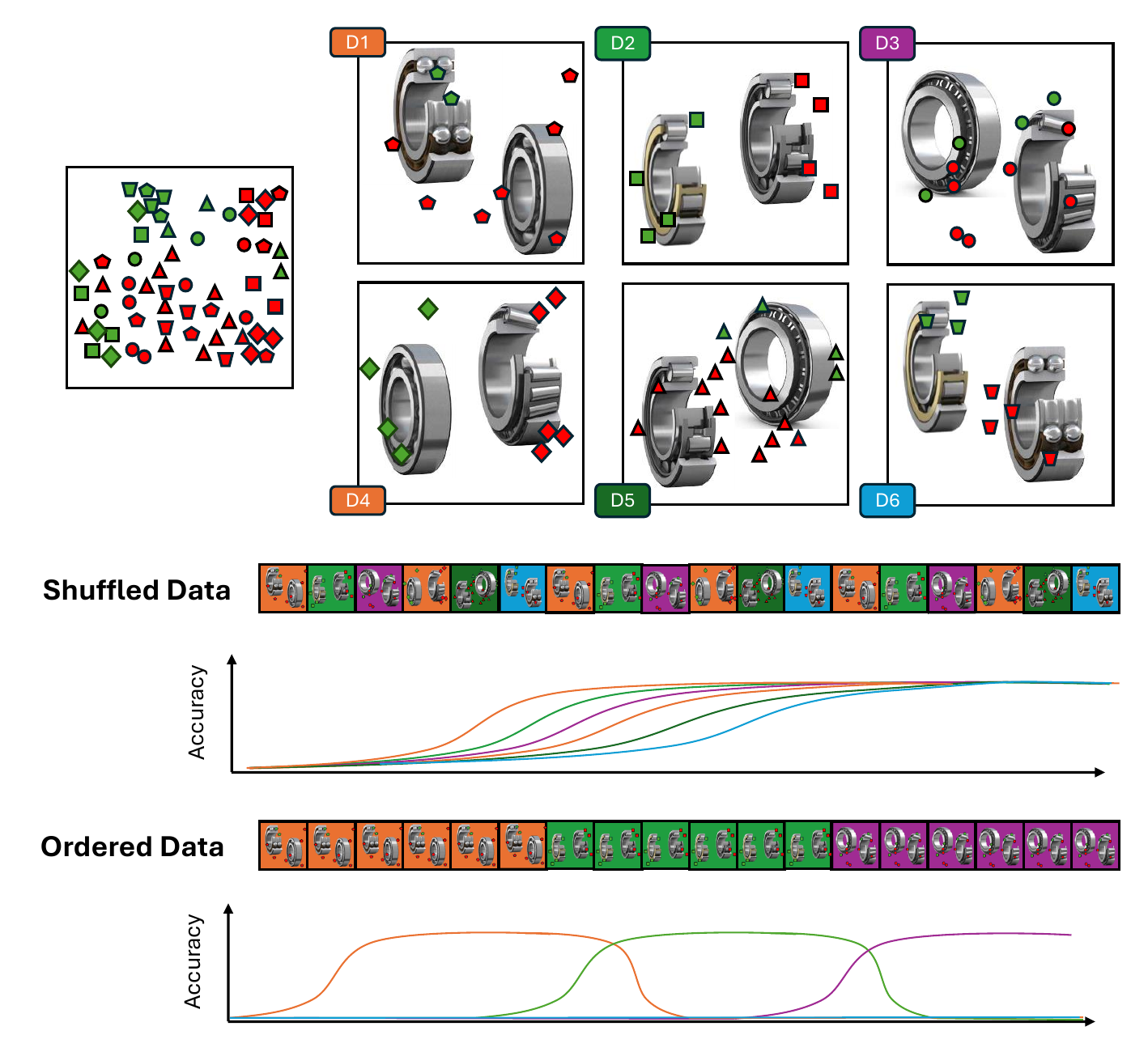}
    \caption{Sequential variation of operational conditions in the data causes catastrophic forgetting}
    \label{fig:traditional_learning}
\end{figure}

\textbf{Traditional data mining.} However, both methods can suffer contamination and distortion from other faults, which can be vulnerable to interference from reflected waves, scattered waves, and mechanical noise radiated from other sound sources. Thus, it becomes harder to locate defective parts of the machine with these methods. In fact, regardless of using vibration or sound signals, it has been shown that the success of fault detection and diagnosis depends on the data mining approach \citep{Heng1997,Karacay2009}. Although these methods can handle the richness of fault types in a static dataset, they are only trained once, built on the assumption that all possible fault types of the target equipment have been completely covered during the training period. However, in industrial scenarios, machines operate under complex and ever-changing circumstances in actual working conditions, with new faults, continuous equipment upgrade, and varying operational conditions appearing at any time during operation, as shown in Figure \ref{fig:traditional_learning}. This challenges the traditional learning paradigm, which assumes the availability of all training data in advance and its independent and identically distributed nature. To avoid this issue, one could fully retrain the designed model. However, this approach not only requires high-capacity storage devices, but also consumes considerable time and effort for model retraining, evidently elevating operation and maintenance costs. Alternatively, one could train the designed model only on the new data. However, since a single model only has access to current data in an individual phase of the learning cycle, it is prone to overfit on the currently available data and suffers performance deterioration on previous data \citep{lin2023theory}.

\textbf{Stability vs. Plasticity.} This phenomenon is often referred to as catastrophic forgetting, and often occurs when learning current data interferes greatly with its ability to apply previous concepts. Ideally, the model should be capable of maintaining concepts seen in the past even long after having learned them, intentionally retiring outdated knowledge only if needed. Naturally, this accumulation of old concept representations is a second-order problem that might undermine performance. It only makes sense if the model will be learning more in the future and concepts share the same structure, e.g., laws of physics underlying the real data, tools developed for similar purposes, data-generating people and organisms with consistent intentions, instead of simply random concepts. In addressing this catastrophic forgetting issue, various continual learning (CL) techniques have been proposed, treating the parameters of neural network models more similar to memory. Thus, ignoring the fact that when the model retains outdated knowledge, it can hinder its plasticity, that is, the ability to adapt fast and effectively learn from new data. A model has lost plasticity if it is unable to optimize its objective function as effectively as a randomly initialized model. Plasticity can thus be thought of as the quality of a particular point in parameter space to serve as a starting point for optimization. The more stages there are in the learning process, the worse this loss \citep{pascanu2021study}.

\textbf{Forward vs. backward transfer.} This stability-plasticity trade-off is related to the challenge of learning invariant representations, i.e. finding a shared solution for all incremental concepts, which risks destroying their adaptability. As more irregular tasks are introduced, the feasible parameter space will tend to narrow, with severe interference between concepts that hurts the knowledge in old and new concepts, a phenomenon known as negative transfer \citep{wang2019characterizing}. Thus, learning all incremental concepts with a shared solution is equivalent to learning each new concept in a limited parameter space that prevents the performance degradation of all old concepts. This problem has proven to be NP-hard in general \citep{knoblauch2020optimal}, because the feasible parameter space tends to be narrow and irregular as more concepts are introduced, thus difficult to identify. Ideally, with incrementally learned concepts being related, it should be possible to exploit their similarity to achieve positive transfer, that is, by learning one concept, the model also becomes better at another concept \citep{lin2023theory}. For example, once a human has learned to play a first musical instrument, it is typically easier for them to master a second one. Ideally, a model should exhibit both forward and backward transfer. Forward transfer consists of learning a concept that facilitates subsequent learned concepts, for example, learning new concepts taking advantage of knowledge extracted from previous concepts and discovering knowledge that might be reusable in the future without knowing what that future might look like. Backward transfer consists of learning a concept that benefits previously learned concepts, i.e., not only avoiding forgetting but also gaining immediate performance in previous concepts which are similar or relevant. In this regard, however, it should be noted that the impact of concept similarity in positive transfer is not monotonic, with intermediate concept similarity being shown to lead to the worst forgetting in the two-concept setup \citep{ramasesh2020anatomy}.

\textbf{This work.} In fact, traditional non-modular CL methods fail to capture the intuition that, in order for knowledge to be maximally reusable, it must capture a self-contained unit that can be composed with similar pieces of knowledge. Building on this research gap, this work proposes a new continual learning method for fault diagnosis, drawing inspiration from gradient boosting, with the widely used CNN as a base learner. While recent work has also drawn inspiration from the AdaBoost algorithm \cite{Schapire2013} to propose a continual learning method based on feature boosting that continuously extends new modules for the initial diagnostic model to fit the residuals between the actual label and its output \citep{he2024new}, this work still maintains a single shared backbone of the fault diagnostic model, lacking in its potential for selective forward and backward transfer. In contrast, this paper presents the first application of modular architectures with knowledge transfer for fault diagnosis in rolling bearings, i.e. allowing combinatorial solutions of previously learned diagnoses to perform new ones. The neurobiological homologue of this approach would be attention: selecting newsworthy information that resolves uncertainty about things you do not already know, given a certain context \citep{vanrullen2021deep}. Experiments performed on a multi-domain fault bearing dataset, measuring average domain accuracy, learning accuracy, and forgetting, while performing statistical tests to assess significant statistical differences, show that this method achieves average domain accuracies in the high 87–89\% range (up to 88.96\% in some configurations), maintains competitive learning accuracies (around 86–87\%), and reduces catastrophic forgetting, with forgetting measures as low as \(2.70 \times 10^{-3}\) in optimal settings. It also shows robust performance across different domain ordering strategies, replay buffer sizes, and even under varying levels of noise corruption. This paper is organized as follows. Section~\ref{sec:related} describes the related work. Section~\ref{sec:multidomain} presents the fault bearing case study, fault types, and corresponding environmental and operational variations. Section~\ref{sec:methodology} presents the continual learning method. Section~\ref{sec:experiments} investigates the role of domain order revealed, performs a quantitative comparison of different domain selection mechanisms for the boosting-inspired experience replay strategy, and studies the model's robustness to noise and data corruption. Finally, Section~\ref{sec:conclusion} covers the conclusion, with avenues for future work.

\section{Related work}
\label{sec:related}

For several years, research has explored the causes, influences, styles, and generating mechanisms of both vibration and acoustic analysis in rolling element bearings \citep{Hawman1988, mba2006development}. In this process, the authors constantly pointed out the limitations of the alternative measurement technique in practical applications. On the one hand, vibration analysis has been accused of having a lack of sensitivity to incipient defects, while acoustic emission is praised for capturing much higher frequencies and minimizing spectral overlap with mechanical vibration signals from rotating machinery \citep{mba2006development, al2006comparative}. Moreover, while acoustic emission is a non-contact method that is easy to set up remotely, vibration sensors are accused of requiring physical contact with the machine, which can be difficult to mount due to irregular machinery geometry, heat damage, and harsh testing environments \citep{li2015incipient}. Conversely, acoustic emission waves through a single-channel microphone are accused of only providing sound pressure values; therefore, being highly sensitive to measurement points, with features of interest being very likely obscured by high-level noise \citep{Lu2019}. Furthermore, despite being a non-contact method, acoustic emission waves are greatly attenuated during propagation. Therefore, like vibration measurement, sensors should be placed as close as possible to the components being tested.

\textbf{Signal processing.} In the early stages, the dominant belief was that the best way to solve this was to meticulously design a signal processing technique exclusively for the task and type of sensor. Firstly, investigating the propagation characteristics in bearings with different operational conditions, such as rotation speeds, radial load, fault-type signals and defect size \citep{Morhain2003}, e.g. via direction-of-arrival estimation \citep{li2023acoustic}, and then process vibration and sound signals, e.g. based on statistical parameters like, crest factor, kurtosis, skewness, beta distribution functions \citep{Heng1997,Karacay2009}, empirical mode decomposition (EMD) \citep{Amarnath2019,li2019application}, envelop spectra \citep{wang2019simple,Amarnath2019,wang2020theoretical}, Wigner-Ville distribution, \citep{baydar2003detection}, wavelet transform \citep{baydar2003detection},  Hilbert--Huang transform \citep{peng2005comparison}, spectral kurtosis \citep{antoni2006spectral}, fast Fourier transform (FFT) \citep{yadav2011audio}, FFT-based nearfield acoustical holography and gray level co-occurrence matrix \citep{lu2012fault,lu2013gearbox}, beamforming and spectral kurtosis \citep{cabada2017fault}, or acoustic imaging and Gabor wavelet transform \citep{wang2018intelligent}.

\textbf{Machine learning.} Over time, as a substantial volume of data has been accumulated on the health status of machinery, and with the increase in artificial intelligence, new data-driven fault diagnosis methodologies emerged. Initially, these signal processing techniques for data feature extraction and selection were paired with traditional machine learning models, e.g. relying on multi-SVM \citep{grebenik2016roller}, and hidden Markov models \citep{jiang2020robust}. However, choosing suitable feature extraction methods always remained a challenging and time-consuming task, because the optimal feature set often varies from case to case in different applications. While the frequency and order of the fault characteristic can be calculated by the geometric parameters and the rotation speeds of the bearings, most of the rolling bearings work under non-stationary conditions, e.g. due to the run-up and shutdown of machines and speed fluctuation of variable loads. When the operational conditions vary, fault characteristics change, e.g. with the spectrum of a nonstationary signal showing a smearing phenomenon. Moreover, even if some methods obtain high-quality time-frequency representations with fine resolution and better energy concentration \citep{Tang2021, Zhao2019} that are effective under speed-varying conditions, prior knowledge is still necessary to calculate the characteristic frequencies and orders of faults.

\textbf{Learned manifold.} Ideally, one should be able to extract information in deployment, learning useful representations from raw data independently of the quality of training data. In this regard, the high degree of automation of the data processing of deep learning (DL) architectures coupled with the increasing size of models and datasets provided a significant step forward, being able to cope with massive data features without expert experience as a foundation. Moreover, over time, it became evident that for certain types of unstructured data, such as time series signals and images, a general-purpose model could be fine-tuned on specific datasets, yielding effective results. Thus, most research efforts pivoted to the multiscale hierarchical latent representations of DL architectures to diagnose faults, instead of manually designed features and shallow algorithms, such as recurrent neural networks (RNNs) \citep{liu2018fault}, deep boltzman machines \citep{li2016gearbox}, deep auto-encoders \citep{zhiyi2020transfer,he2019improved}, deep belief networks \citep{tang2018adaptive}, multilayer spiking neural networks \citep{zuo2022multi}, Bayesian deep learning \citep{zhou2022towards}, generative adversarial networks \citep{dai2023categorical,cao2022method,shao2023dual}, capsule networks \citep{Yan2022}, or even hybrid neural networks with principal component analysis \citep{you2022rolling}.

\textbf{Convolution}. Despite this wide array of DL methods, the most widely used method remains convolutional neural networks (CNN) \citep{Chen2019,Li2020a,yao2018end,Lei2022,Wang2022}, enhanced with dynamic training rates \citep{Guo2020}, time series transformer \citep{lu2023rotating}, or hidden Markov models \citep{wang2018convolutional}. As an input, these rely on wavelet transform \citep{Shao2020}, acoustic images reconstructed from the acoustic field of a microphone with the wave superposition method \citep{9144221}, frequency spectra of vibration signals \citep{janssens2018deep}, wavelet packet energy image as input for spindle bearing fault diagnosis \citep{ding2017energy}, 2-D and 3-D conversions of one-dimensional vibration time series \citep{wen2018new,li2023cnn}, maps of cyclic spectral coherence \citep{chen2020deep}, Fast Fourier Transform (FFT) \citep{Chen_2024} and Markov transition field \citep{Chen_2024,Lei2022,Wang2022,Yan2022}. Alternatively, one can also use a one-dimensional deep CNN, which can effectively learn discriminative features from raw signals \citep{dong2023intelligent,huang2019deep}. Furthermore, this ability to process multidimensional data enabled the fusion of heterogeneous monitoring signals. For example, using domain knowledge, operating conditions, and vibration fused into a three-dimensional input \citep{Guo2020}, extracting multiple source domains with time-varying working conditions \citep{wang2022multisource,gao2024multi}, using raw data from horizontal and vertical vibration signals \citep{Chen2019}, combining both vibration signals and current signals \citep{Shao2020}, infrared thermal images and vibration signals \citep{miao2024deep}, or multichannel information from sensors at different locations \citep{yao2018end,Li2020a}. 

\textbf{Continual learning.} Furthermore, proper adaptation in changing environments requires not only parameter adaptation, but also structural expansion in an incremental manner. Nonetheless, the current literature on fault diagnosis focused exclusively on stability to prevent forgetting the knowledge, disregarding the plasticity the model needs to adapt to new knowledge. On one hand, these techniques focus in controlling how model parameters change between concepts so that there are independent representations for each concept. For example, regularization-based core space gradient projection guide gradient descent along the orthogonal direction of the previously input subspace \citep{ren2024core}, or dynamic weight correction to fine-tune the model’s response to new tasks \citep{li2023deep,hu2024adaptive}. Adaptive feature consolidation residual networks consolidate important features for previously learned tasks, which helps retain performance on past tasks, while adapting feature representations to accommodate new tasks, typically through re-weighting or adjusting internal parameters, rather than expanding the model \citep{zhang2024adaptive}. Feature-based knowledge distillation consists of transferring the feature representations or intermediate activations of the teacher model for each task to a smaller fixed capacity model student model \citep{chen2023continual,min2023incremental}. On the other hand, some methods focus on capturing a common structure within various tasks with an aggregated state abstraction. For example, with an incremental multitask shared classifier that adds new output heads for each task, while the core of the architecture that allows shared learning remains fixed in size \citep{wang2023bearing}. Alternatively, a dual-branch aggregated residual networks allows one to keep one branch that maintains representations of previous tasks, helping to prevent catastrophic forgetting, while the other branch is adaptive, allowing it to learn new features as new tasks are introduced \citep{chen2022lifelong,chen2023continual}. Finally, some methods focus on replay-based techniques to retain task knowledge without expanding the architecture, for example, through the distribution projection replay module \citep{zhou2024distribution}, generative feature replay \citep{liu2023lifelong}, or repetitive replay with memory indexing \citep{zheng2022bearing}.

\section{Multi-domain rolling bearing}
\label{sec:multidomain}

Although real-world data are key to assess any data mining approach, all surveyed datasets lack the necessary diversity, contextual information, and time-stamps to properly validate a continual learning method. To avoid this issue, simulated multi-domain data provides the ability to conduct controlled and repeatable experiments, enabling researchers to manipulate variations in environmental and operational conditions and observe their impacts without the constraints of real-world data collection.

\begin{figure}[ht]
    \centering
    \includegraphics[width=0.3\textwidth]{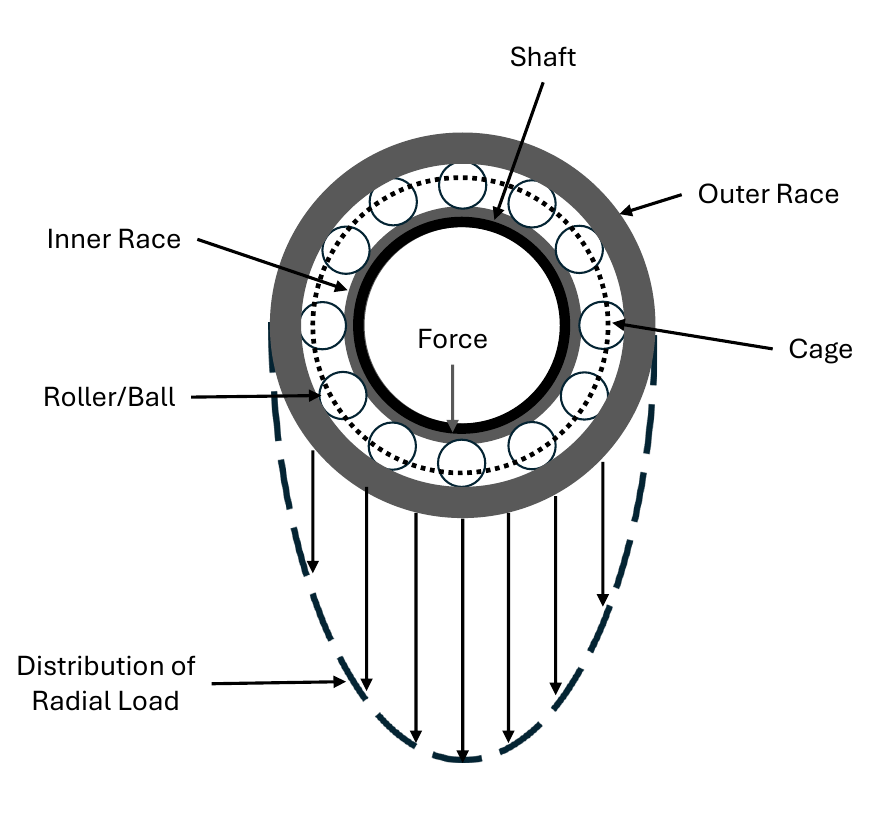}
    \caption{Rolling bearing in radial load condition}
    \label{fig:a}
\end{figure}

\textbf{EOV requirements.} Ideally, domains in a dataset should represent unique configurations or combinations of these conditions, capturing the sequential changes typical of real-world industrial settings. For instance, varying load types, with radial, axial, and combined radial/axial loads, to replicate the range of mechanical stresses bearings encounter across different applications, as illustrated in Figure \ref{fig:a}. Rotational speeds should also vary, encompassing low, medium, and high speed levels as defined by production requirements. As the bearing operates at different speeds and loads, the number of rollers and their positions in the loading zone change with the angular positions of the shaft, resulting in periodic variations in support stiffness. In addition, the dataset should also reflect diverse measurement techniques to account for variations in data collection. Data acquisition should ideally span multiple sampling rates to mimic different sensor configurations and potential resource limitations. Sensor types and positioning, both in terms of orientation and location relative to the bearing, should be diversified to capture variability in signal quality. In terms of rolling bearing structure, the dataset should include different types of bearings, such as plain, needle, cylindrical, and magnetic bearings, each embedded in various types of rotating machinery. Finally, the dataset should introduce secondary component conditions, such as misalignment, imbalance, and looseness, which indirectly affect bearing health. 

\textbf{Fault requirements.} Within each domain, the diagnosis of bearing faults targets various components, such as the outer race, the inner race, the rolling elements, and the cage, each representing different sources in faults, as shown in Figure~\ref{fig:a}. Although fatigue cracking is the most prevalent failure type, arising primarily from high stress in heavily loaded contact zones typical of radial load conditions, failure modes can vary significantly. Extended operating periods and increased fatigue load cycles often lead to localized faults, including pits and spalls on rolling contact surfaces. These defects not only affect the bearing surface, but may also alter the lubrication system, a critical factor in the maintenance on effective operating conditions. The development of faults such as micro-pitting, macro-pitting, and spalling, with defect sizes ranging from 20 to 100 $\mu$, can push the bearing into sub-optimal or abnormal operating states. Spalling, for instance, initiates as micron-scale imperfections that gradually extend across the raceway, creating larger zones of damage. As the defect area grows, it can impact the rolling elements, leading to spall formation on the mating balls or rollers. This wear damage is then transferred to opposing surfaces, accelerating degradation. In bearings, high stress concentrations at points or along lines of contact are strongly correlated with spalling, underscoring the need for continuous monitoring and early fault detection in high-stress applications.

\textbf{Available datasets.} Although some studies proposed different simulated datasets, most of these don't have the necessary diversity to validate a truly complex multi-domain environment with the interconnected dynamics of real-world machinery. Most datasets focus exclusively on a single source of non-stationary data, for example, varying rotating speed \citep{huang2018bearing,lessmeier2016condition,marshall2023dataset}, and different bearing models \citep{sehri2023university,thuan2023hust}. To overcome these limitations, a recent dataset provides data collected for three types of bearing faults, with known contextual information of a diverse range of environmental and operational conditions, for example, varying types of bearings, sampling rates, types of conditioning on environmental rolling components and rotation speeds \citep{lee2024multi}.

\textbf{Dataset description.} Using this dataset, a multi-domain partition was created, as will be described. We provide here all the minimum information necessary to understand the dataset, with further details available elsewhere \citep{lee2024multi}. For data collection, a PCB Piezotronics 333D01 accelerometer, with sensitivity 4.00 \% FSV/g and measurement range of \( \pm 20 \) g, was mounted with a magnetic stud on the top side of the bearing housing at the shaft end, with two rotor disks for a brushless DC motor has a 40 W power output, a 60 Hz frequency, and a maximum rotating speed of 1700 RPM. All instances were transformed from time series signals into 2D representations to leverage the capabilities of convolutional neural networks (CNNs), a widely adopted approach in rolling bearing fault diagnosis. For this purpose, the Markov transition field (MTF) spectrogram was used to encode the signal dynamics into structured images, following recent findings that indicate MTF effectively maintains temporal relationships and offers a comprehensive depiction of state transitions within the data, making it highly suitable for identifying subtle discrepancies required in fault detection \citep{Yan2022, Chen_2024, Lei2022, Wang2022}. Table~\ref{tab:domains} provides a summary of the 18 resulting domains, each consisting of 7200 instances.

\begin{table}[ht]
    \centering
    \caption{Domains by bearing, faults, environment and speed. B = ball, IR = inner-raceway, OR = outer-raceway, H = healthy, L = looseness, U = unbalance, M = misalignment, S = Slow (600, 800, 1000 RPM), F = Fast (1200, 1400, 1600 RPM)}
    \label{tab:domains}
    \begin{tabular}{@{}cp{0.9cm}p{1.8cm}p{2.1cm}c@{}}
    \toprule
    \textbf{Domain} & \textbf{Bearing} & \textbf{Faults} & \textbf{Environment} & \textbf{Speed} \\
    \midrule
    1/2     & \multirow{3}{*}{6204}        & \multirow{3}{=}{B, IR, OR, H} & H, M1, U1, L       & \multirow{3}{*}{S / F} \\
    3/4     &                               &                                & H, U1, U2, U3      &  \\
    5/6     &                               &                                & H, M1, M2, M3      &  \\
    \cmidrule(l){1-5}
    7/8     & \multirow{3}{*}{30204}       & \multirow{3}{=}{B, IR, OR, H} & H, M1, U1, L       & \multirow{3}{*}{S / F} \\
    9/10    &                               &                                & H, U1, U2, U3      & \\
    11/12   &                               &                                & H, M1, M2, M3      &  \\
    \cmidrule(l){1-5}
    13/14   & \multirow{3}{*}{N(J)204} & \multirow{3}{=}{OR, H, IR}    & H, M1, U1, L       & \multirow{3}{*}{S / F} \\
    15/16   &                               &                                & H, U1, U2, U3      &  \\
    17/18   &                               &                                & H, M1, M2, M3      &  \\
    \bottomrule
    \end{tabular}
\end{table}

\textbf{Bearing types.} Firstly, three types of bearings, differing in their design and load handling capabilities, were used to create multi-domain data: deep groove ball bearings (model 6204), cylindrical roller bearings (models N204 and NJ204), and tapered roller bearings (model 30204). Tapered roller bearings feature conical rollers and are designed to handle radial and axial loads efficiently. The tapered design allows these bearings to accommodate combined loads, making them ideal for applications involving higher load conditions, such as in automotive or heavy machinery. Deep groove ball bearings are the most common bearing type, consisting of an inner and outer ring with a set of balls between them. These are versatile and designed to handle both radial and axial loads in both directions, making them suitable for general applications. Cylindrical roller bearings, on the other hand, use cylindrical rollers instead of balls, which increases the contact area with the raceways, allowing them to handle higher radial loads with reduced wear and increased load capacity. However, they are less capable of managing axial loads compared to ball bearings. Moreover, to represent two types of cylindrical roller, models were used, in which the main difference lies in the N204 bearing being able to separate its outer raceway, while the NJ204 bearing can separate its inner raceway.

\begin{figure}[!ht]
    \centering
    \includegraphics[width=0.45\textwidth]{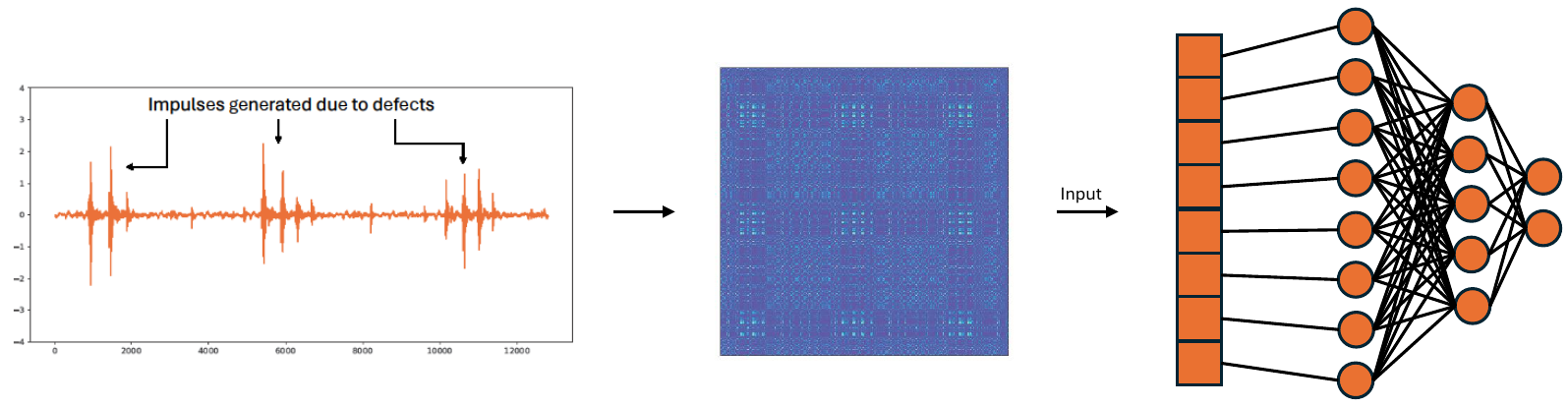}
    \caption{Signal processing with Markov transition field}
    \label{fig:mtfsignal}
\end{figure}

\textbf{Fault types.} Secondly, three different types of rotating component faults were introduced: looseness (L), unbalance (U), and misalignment (M). The L fault was caused by loosening the bearing housing screw at the motor end by half a turn. For M faults, which were classified into three severity levels (1, 2, and 3), the central axis of the BLDC motor was displaced by 0.6, 0.8 and 1.0 mm, respectively. The U fault was induced by adding an additional screw to the rotor disk, with the mass added at three severity levels (1, 2, and 3) corresponding to 3 g, 4 g, and 5 g, respectively. In both cases, higher numbers correspond to greater fault severity.

\textbf{Environments and speed.} Thirdly, two distinct ranges of rotational speeds were considered: low speeds of 600, 800, and 1000 RPM, and high speeds of 1200, 1400, and 1600 RPM. Furthermore, in all domains, variations can be observed in the distribution of bearing fault locations, sampling rates, and noise levels. Bearing fault locations include ball (B), inner-raceway (IR), and outer-raceway (OR) faults, with defects manufactured using the grinding method. In terms of sampling rates, data was recorded using an accelerometer for 160 seconds at 8 kHz and for 80 seconds at 16 kHz, ensuring an equal number of data points. In addition, a noise level of 5 \% was introduced in all instances to increase the number of instances, while simulating realistic measurement conditions.

\section{Methodology}
\label{sec:methodology}

\begin{figure}[!ht]
    \centering
    \includegraphics[width=0.5\textwidth]{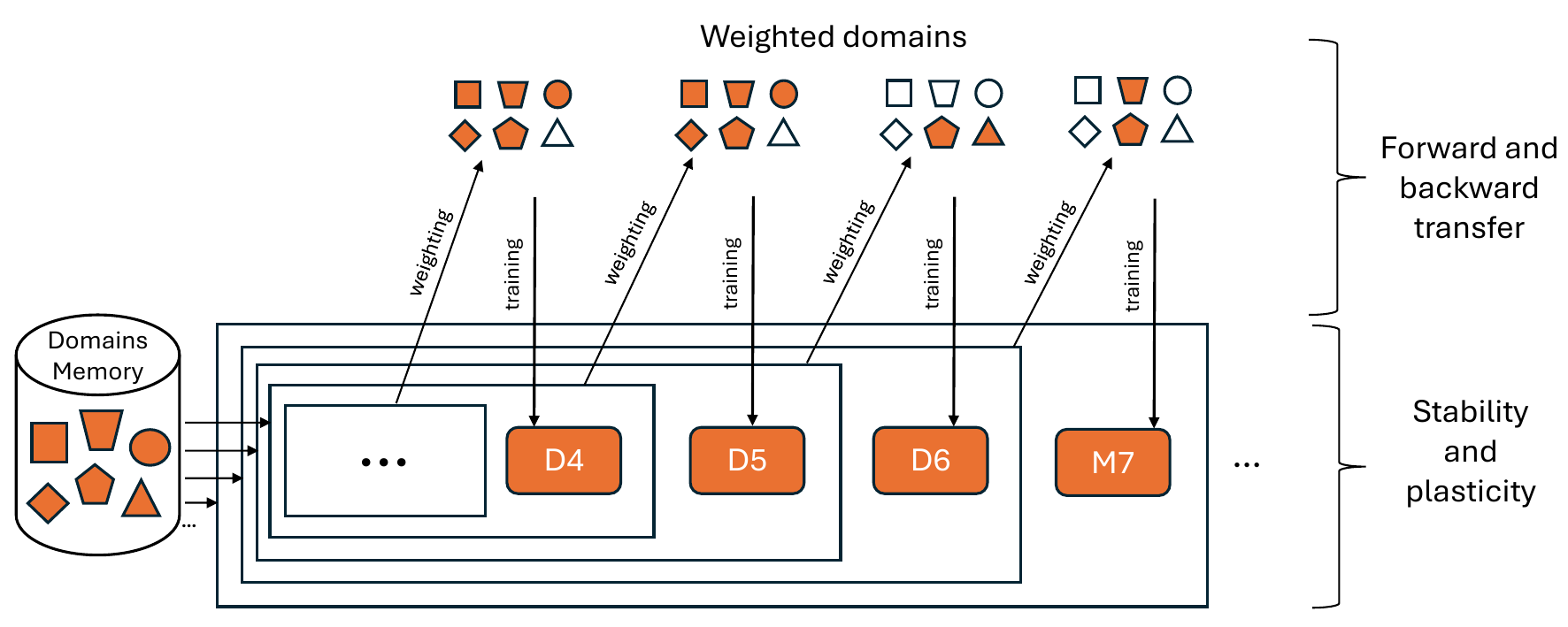}
    \caption{Boosting-inspired modular ensemble CNN architecture for cross-domain learning}
    \label{fig:boosting}
\end{figure}

To describe the proposed fault diagnosis methodology, the various algorithmic components developed on top of a traditional two-layer CNN are presented in light of the four second-order requirements that one must consider in a continual learning setting: catastrophic forgetting, where the model is prone to overfitting on the currently available data and suffers from performance deterioration on previous data; lack of plasticity, where the model holds on to outdated knowledge, losing the ability to effectively learn, adapt fast, and generalize from new data; leveraging forward transfer, in which learning a new concept takes advantage of knowledge extracted from previous concepts, as well as discovering knowledge that might be reusable in the future; and leveraging backward transfer, in which learning a concept benefits previously learned concepts. The initial stage of the approach involves a feature generator and isolated domain-specific classifiers that allow for a continually growing capacity as more domains emerge, ensuring that the models do not interfere with each other's learning and retaining plasticity. In order to mitigate the risk of the model forgetting earlier domains while adapting to new ones, a restricted experience replay mechanism was developed. The second stage of the approach focused on leveraging on the forward and backward transfer opportunities of nonlinear environmental and operational influences by selectively choosing which domains to use for training models sequentially, such that each new model incorporates knowledge from the domains with the highest error in the previous episode. Figure \ref{fig:boosting} provides an overview of the approach.

\subsection{Stability and plasticity}

The model architecture is operationalized to enable growth capacity in a sequence of domains, while sharing knowledge from previous domains. Assume that a sequence of domains $D_1, \ldots, D_n$ is presented to the system, each sharing the same input $X$, but different outputs $Y_1, \ldots, Y_n$. In each episode $k$, the model is tasked with training in the current domain $D_k$ and a selected subset of previous domains to facilitate knowledge sharing. For example, during episode $k = 2$, the training involves a feature generator $h$ and domain-specific classifiers, leading to the formation of the models $g_1$ $ \circ $ $h : X \rightarrow Y_1$ and $g_2$ $ \circ $ $h : X \rightarrow Y_2$. The model then classifies the inputs from both domains, producing a probability vector \( p_{g_i \circ h}(y | x), \forall y \in Y_i \) based on the respective domain. In episode \(k\), such set of domains can be defined as \(\overline{D}_k = \{D_{w_k^1}, \ldots, D_{w_k^b}\}\), where \(b \leq k\) serves as a hyper parameter, and \(\omega_k^i \in \{1, \ldots, k\}\). Training in \(\overline{D}_k\) involves the use of a feature generator $h_k$ and domain-specific classifiers \( g_{(k, \omega_k^i)} \) for each chosen domain. 

\begin{figure}[!ht]
    \centering
    \includegraphics[width=0.5\textwidth]{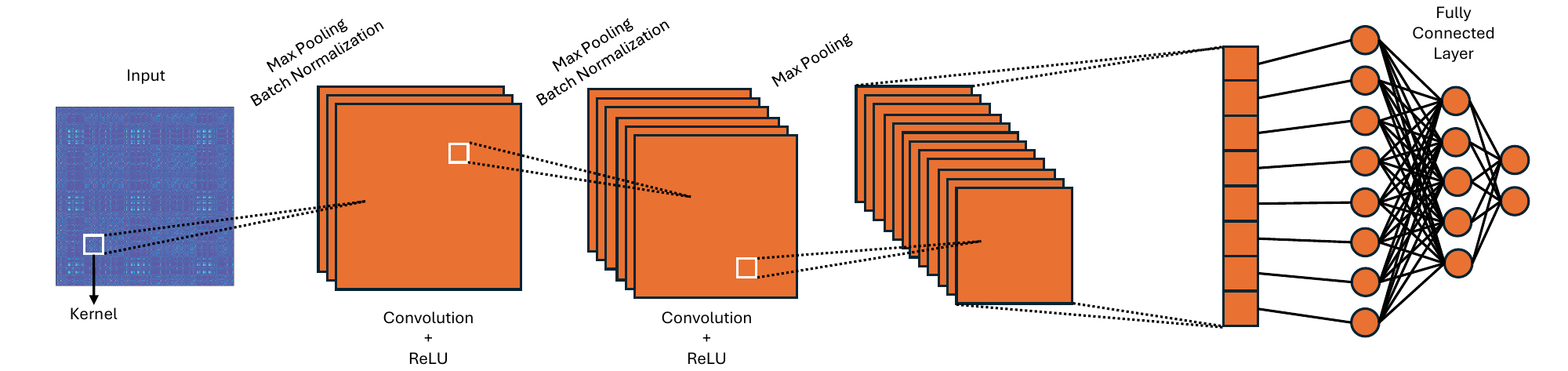}
    \caption{MTF through layers of convolution and pooling}
    \label{fig:layers}
\end{figure}

\textbf{Feature generator.} The shared feature generator supports the domain-specific CNNs, pulling out meaningful features from the input data, acting not only as a filter but also enhancing essential details of the input. Moreover, this centralized feature extraction process is extremely cost-effective for continual learning with unbounded data streams, while at the same time allowing each classifier to specialize in its designated domain \cite{Tripuraneni2020}. In practice, the implemented feature generator was composed of two convolutional layers with 80 filters and a kernel size of 3x3 pixels to detect important features related to a potential anomaly, as shown in Figure ~\ref{fig:layers}. Each of these convolutional layers applies a set of filters, which can be thought of as small sliding windows that move over the image to detect different patterns, followed by a max pooling operation that down-samples the dimensionality of the data \citep{Krizhevsky2017}. To ensure that the learning process is stable and effective, the model employs Batch Normalization \citep{Ioffe2015}. The abstract features extracted from these shared convolutional and pooling layers are subsequently fed into a fully connected layer, which is structured to manage multiple domains.

\textbf{Domain-specific classifiers.} Unlike traditional models that try to handle all domains with a single larger model, this model divides domains into several smaller CNNs, with each one being trained in a specific domain or in a group of specific domains. Such process is akin to using an ensemble of smaller CNNs. This ensemble not only improves robustness, but also boosts the overall performance thanks to a higher diversity, heterogeneity, and de-correlated predictions \cite{havasi2020training}. Indeed, while their functionality in continual learning scenarios has only recently been fully studied, such benefits have been well known for supervised learning \cite{fort2019deep}. Not only modular adaptation of individual models leads to an attenuation of forgetting and a boost in the overall performance by ensuring that models do not interfere with each other's learning \cite{caccia2022anytime}, but also can help reduce extra parameter costs for task-specific sub-networks \cite{wen2020batchensemble} and save computational cost \cite{doan2023continual}, by dividing the workload. Ultimately, these domain-specific models collectively form the current model, with the ability to predict data from $D_i$ for \(i \leq k\) being derived from averaging class probabilities output by all models that were applied to that domain:

\begin{equation}
    p_{k,i}(y|x) \propto \sum_{l=1}^{k} 1_{\{P_i \in \overline{P}_l\}} g_{l,i} \circ h_l(x)
    \label{dataprediction}
\end{equation}

\subsection{Forward and backward transfer}

With these domain-specific classifiers, the key challenge lies in reformulating the forgetting problem into a task interference problem and solve it using model selection to discover cooperative domains. For this purpose, one can implicitly differentiate helpful and harmful knowledge based on the structural allocation obtained from the disjoint subset of parameters to each domain allows to not suffer from updating old knowledge with new one \cite{mallya2018packnet,serra2018overcoming,abati2020conditional}. However, such approach still lacks a way to guide the search for relationships between domains, e.g. selecting the optimal domain-classifier based on the similarity of the Gaussian distributions of each class \cite{rypesc2024divide}. For this purpose, this work follows the idea of introducing sensitivity measures to the loss of the current domain from the associated domains to find cooperative relations \cite{jin2022helpful}, by emulating the boosting process for selecting domains to train with \cite{Ramesh2022}.

\begin{figure}[!ht]
    \centering
    \includegraphics[width=0.5\textwidth]{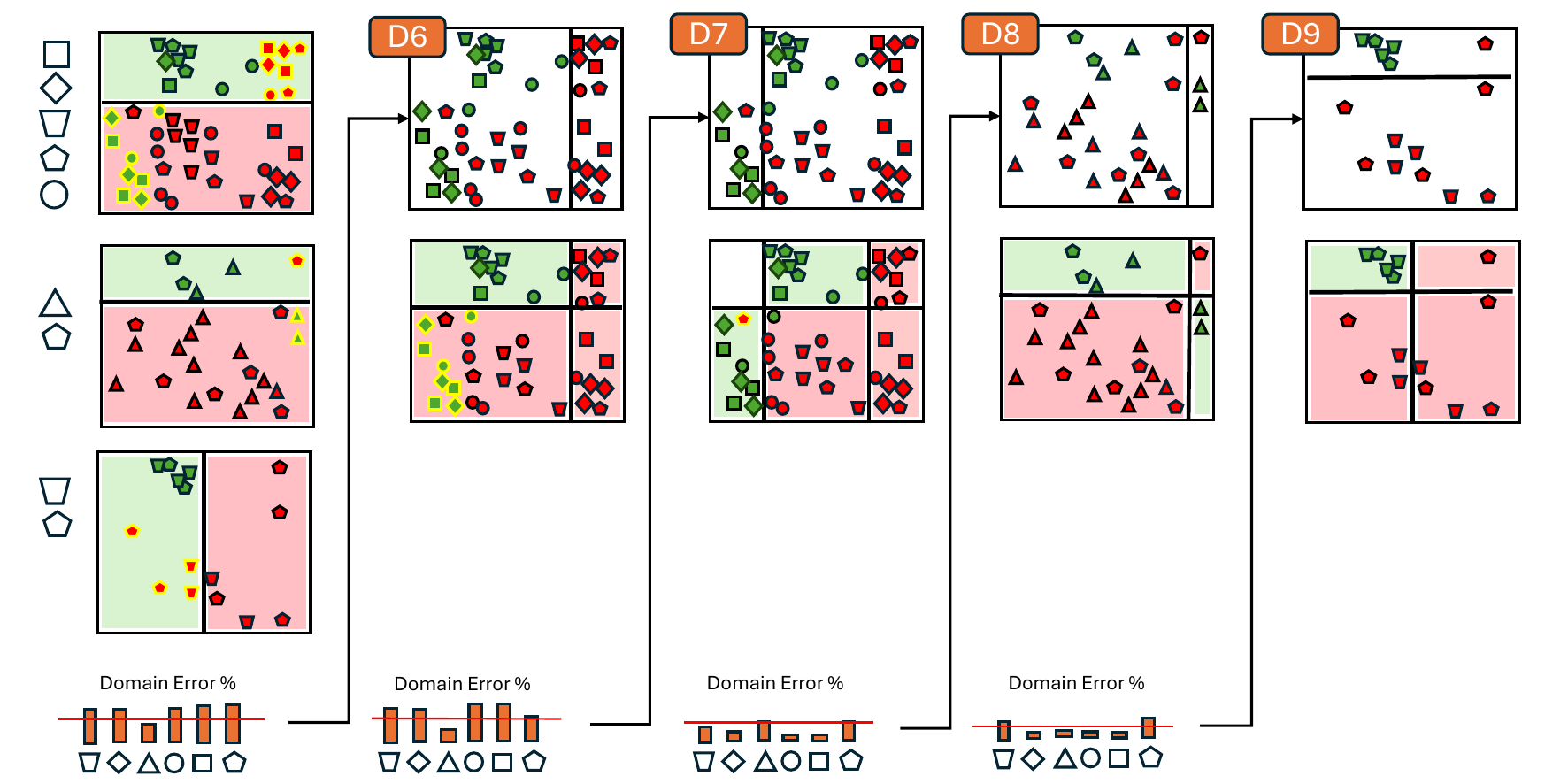}
    \caption{Domain selection and error-driven retraining process across episodes}
    \label{fig:training_process}
\end{figure}

\textbf{Boosting-inspired transfer.} In the traditional Adaboost \cite{Schapire2013}, the training weights for each instance in the next episode are adjusted on the basis of the performance weaknesses of each individual model. Conversely, in this approach, the weights for the next training episode are based on the performance of the entire ensemble up to that point, not just individual CNNs. This difference allows the model to adapt its learning more effectively across multiple domains considering the collective knowledge of the ensemble. New models are trained sequentially, with each new model incorporating knowledge from the domains with the highest error in the previous episode. After each domain is learned, the system introduces a new model that focuses on the domains with the greatest need for improvement, identified by their error rates. This difference allows the model to adapt its learning more effectively across multiple domains considering the collective knowledge of the ensemble. Assuming that $\overline{w}_{k,i} \in \mathbb{R}^n$ is a normalized vector of domain-specific weights, after episode \(k\):
\begin{equation}
    \overline{w}_{k,i} \propto \exp\left(-1/m \sum\nolimits_{(x,y) \in S_i} \log p_{k,i}(y | x)\right)
\end{equation}
for each domain \(D_i\) with \(i \leq k\); for \(i > k, \overline{w}_{k,i} = 0\). Subsequently, in the following episode, domains \(\overline{D}_{k+1}\) are drawn from a multinomial distribution with weights \(\overline{w}_{k}\). With this, it makes it possible to put lower weight on domains with a lower empirical risk for the next boosting episode. Thus, ensuring the system progressively concentrates on harder-to-classify domains, similarly to how AdaBoost reduces the training error by progressively focusing on difficult samples \cite{Schapire2013}. Figure \ref{fig:training_process} illustrates this process, where at each step, domains are evaluated based on their error percentages, with domains exceeding the error threshold prioritized for retraining in the subsequent episode.

\begin{figure}[!ht]
    \centering
    \includegraphics[width=0.5\textwidth]{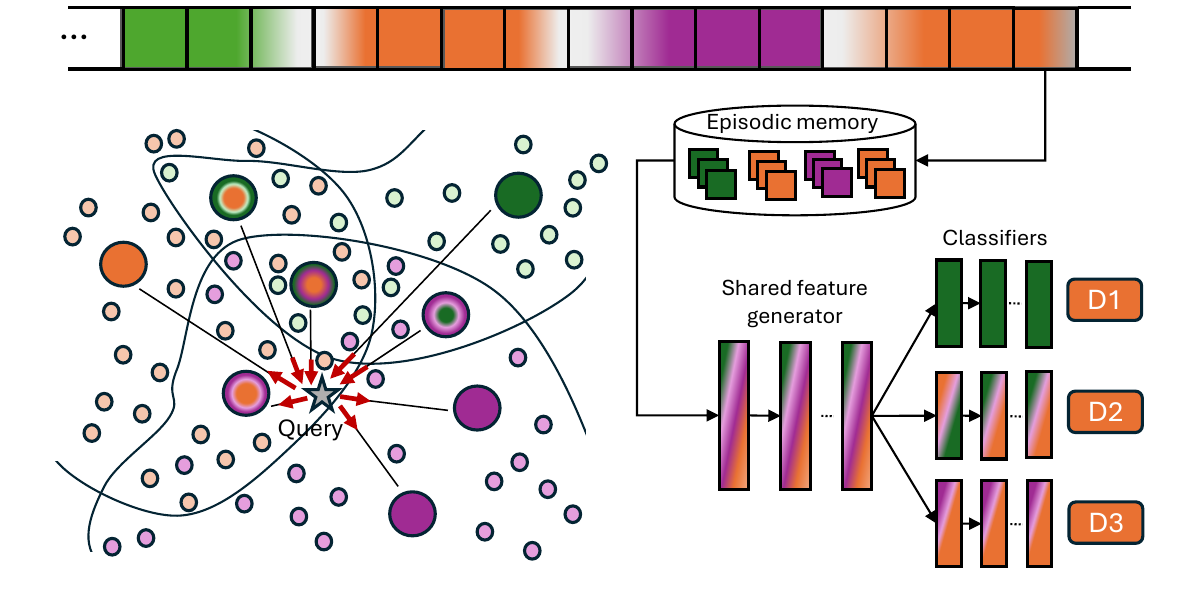}
    \caption{Knowledge sharing between domains}
    \label{fig:stabilitycapacity}
\end{figure}

\textbf{Experience replay.} Naturally, this training process requires revisiting a small fraction of data from previous domains that are picked to retrain with. For this purpose, the architecture integrates a restricted data replay mechanism, that stores only 10\% of the data from the past domains. For the selected domains in each iteration, stored samples are combined uniformly across each mini-batch to contain an equal number of samples from all past and current domains. However, while traditional experience replay is normally used to created shared invariant solutions, in this case each domain-specific classifier selectively receives different replayed domains. This grouping allows for controllability in knowledge sharing between these groups, for a more coordinated adaptation to challenging domains, progressively concentrating on harder-to-classify domains by leveraging the collective knowledge of the ensemble. As represented in Figure \ref{fig:stabilitycapacity}, each domain shares information with others, allowing new domains to benefit not only from internal knowledge but also from the insights gained by overlapping or adjacent domains. Thus, facilitating smoother transitions and faster learning when encountering new operational conditions. Related domains can use their similarity for positive transfer, where learning one domain enhances performance in another or simplifies its (re)learning. Such transfer can be observed forward, with an old domain aiding current domains, or backwards, with current domains benefiting previously learned ones \cite{lopezpaz2017}. Assume that $\overline{w}_{k,i} \in \mathbb{R}^n$ is a normalized vector of domain-specific weights, after episode \(k\), domains \(\overline{D}_{k+1}\) are drawn from a multinomial distribution with weights \(\overline{w}_{k}\) using a lower weight for domains with a lower empirical risk in the previous boosting episode:
\[
\overline{w}_{k,i} \propto \exp\left(-1/m \sum\nolimits_{(x,y) \in S_i} \log p_{k,i}(y | x)\right)
\]

\section{Experiments}
\label{sec:experiments}

To evaluate the effectiveness of the proposed methodology for continual learning, three sets of quantitative experiments were performed. Firstly, investigating the role of domain order and the effect of curriculum design on continual learning performance, instead of relying on classes introduced sequentially based on arbitrary criteria. Secondly, providing a quantitative comparison of domain selection mechanisms for the boosting-inspired experience replay strategy. Thirdly, evaluating the model's robustness to noise and corruption commonly encountered in real-world applications. These experiments implicitly or explicitly address the four previously formulated requirements of catastrophic forgetting, lack of plasticity, forward transfer, and backward transfer. Thus, all experiments refer to three evaluation metrics: the average domain accuracy (\(ACC\)), learning accuracy (\(LA\)) or forgetting measure (\(FM\)). \(ACC\) evaluates the overall performance of the model in all domains, i.e. $\frac{1}{D} \sum_{i=1}^{D} a_{D,i}$, where \(a_{D,i}\) represents the accuracy on the \(i^{th}\) domain after training all the \(D\) domains, and \(D\) is the total number of domains. Higher \(ACC\) values indicate better final performance across all domains. \(LA\) evaluates the model's ability to learn new domains by using prior knowledge, i.e. $\frac{1}{D} \sum_{i=1}^{D} a_{i,i}$, where \(a_{i,i}\) represents the accuracy immediately after training in the \(i^{th}\) domain. Higher values indicate better learning transfer across domains. \(FM\) evaluates how much the model has forgotten previous domains after learning new domains, i.e. $\frac{1}{D - 1} \sum_{i=1}^{D-1} \max_{l \in \{1, \ldots, D-1\}} (a_{l,i} - a_{D,i})$, where \(a_{l,i}\) represents the accuracy in the \(i^{th}\) domain after learning the \(l^{th}\) domain, and \(a_{D,i}\) represents the accuracy in the \(i^{th}\) domain after learning all domains. Lower values are better, indicating that the model retains more knowledge from previous domains. All results are presented over 10 seeds.

\begin{figure}[!ht]
    \centering
    \includegraphics[width=0.35\textwidth]{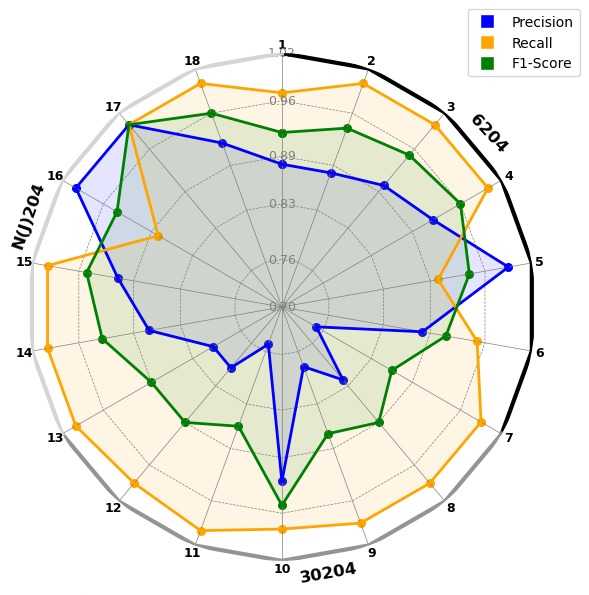}
    \caption{Performance metrics for each bearing and domain}
    \label{fig:spider}
\end{figure}

\begin{figure}[!ht]
    \centering
    \includegraphics[width=0.48\textwidth]{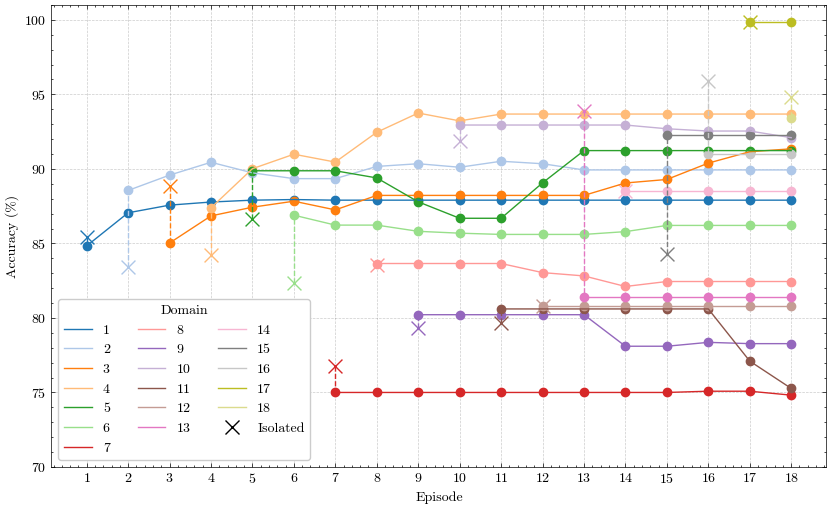}
    \caption{Baseline model accuracy}
    \label{fig:baseline}
\end{figure}

\textbf{Baseline.} The baseline model serves as a reference point for comparison in all experiments. It represents a standard continual learning approach without incorporating domain ordering, exemplar selection, or corruption-specific mechanisms. The baseline achieves an average \(ACC\) of \(86.35\%\), a \(LA\) of \(85.87\%\), and a \(FM\) of \(5.07 \times 10^{-3}\). Domain-specific metrics for the baseline, illustrated in Figure~\ref{fig:spider}, reveal consistent challenges in certain domains, particularly domains 7 and 11, which exhibit lower precision, recall, and F1-scores compared to other domains. These observations highlight inherent difficulties in these domains that persist under different experimental conditions. Moreover, the performance of the baseline model is shown in Figure~\ref{fig:baseline}, which shows the accuracy in the 18 domains. Domains 7 and 11 consistently perform poorly compared to other domains in all experiments. Specifically, domain 7 achieves an F1-score of only 0.86 despite perfect recall (1.00), indicating poor precision (0.75). Domain 11 exhibits similar challenges with an F1-score of 0.86 due to low precision (0.75), despite achieving perfect recall (1.00). These results highlight inherent difficulties in these domains, likely due to unique characteristics or noise within their data. Although the baseline model demonstrates resilience to mild noise levels, its performance deteriorates with higher levels of corruption (\(ACC\) = 86.35\%). These findings emphasize the importance of designing robust mechanisms to handle challenging domains and noisy environments effectively. Lower \(FM\) values across experiments highlighted the effective retention of past knowledge while maintaining competitive accuracy levels across new tasks.

\begin{figure}[!ht]
    \centering
    \includegraphics[width=0.48\textwidth]{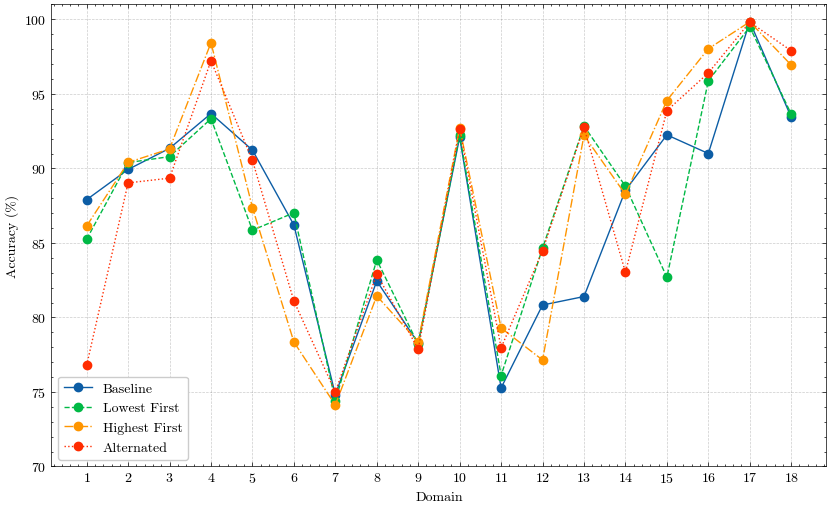}
    \caption{Accuracy with different domain training orders}
    \label{fig:domain_ordering}
\end{figure}

\textbf{Domain ordering.} Learning curricula can play a crucial role in continual learning, significantly impacting the model's ability to learn new tasks while retaining existing knowledge. In this regard, the principles in learning a curriculum suggest that learning is more effective when the examples progress from simple to complex, mirroring how humans and animals learn \cite{mannekote2024}. However, it has been shown this easy to hard strategy is not always ideal, and reversing the difficulty ranking from hard to easy can also achieve the best performance \citep{zhang2018empirical,shrivastava2016training}. Moreover, CNNs have been found to derive most learning values from the hardest examples, and the damage of excluding those easiest examples is minor \citep{avramova2015curriculum}, or can even hurt performance and delay learning \citep{hacohen2019power}.  Thus, three distinct methodologies are performed for sequencing the domains based on their baseline accuracy: lowest first, highest first and alternated. The lowest first strategy begins with the domain showing the lowest accuracy and progresses through domains with progressively higher accuracy levels. Thus, potentially helping the model learn from its mistakes early on. Conversely, the highest first strategy trains in the order of gradually decreasing accuracy levels, potentially helping the model rapidly acquire a diverse set of representations, while preventing the classifier from experiencing immediate confusion from similar tasks. Finally, the alternated strategy cycles through domains of highest to lowest accuracy, which may help maintain the balance between reinforcing strengths and addressing weaknesses. Figure~\ref{fig:domain_ordering} illustrates the accuracy in the domains for each domain ordering strategy, showing how the different sequencing methods impact the model performance as the training progresses through the domains. To provide a more detailed comparison in terms of forward and backward transfer, Table~\ref{tab:domain_ordering} summarizes the \(ACC\), \(LA\), e \(FM\) for each domain order strategy. As shown the strategy that uses the domains with the highest accuracy achieves the highest average accuracy (\(ACC = 88.05\%\)), suggesting that starting with the high-performing domains allows the model to establish a strong foundation for subsequent learning. However, this strategy also results in a slightly higher forgetting measure (\(FM = 8.23 \times 10^{-3}\)), indicating that prioritizing high-performing domains could lead to greater catastrophic forgetting in earlier domains. Interestingly, the alternate strategy achieves the lowest forgetting measure (\(FM = 3.65 \times 10^{-3}\)), demonstrating its effectiveness in mitigating catastrophic forgetting by alternating between high- and low-performance domains during training episodes. This strategy balances retention across domains while maintaining competitive accuracy levels (\(ACC = 87.70\%\)). Furthermore, domain-specific observations highlight challenges in certain domains, particularly domains 7 and 11. Domain 7 consistently shows a lower accuracy (\(\approx 75\%\)) in all strategies, indicating inherent difficulty or a lack of representational overlap with other domains. Similarly, domain 11 shows a significant performance drop compared to other domains, suggesting sensitivity to domain ordering or potential issues with data quality or feature representation.

\begin{table}[!ht]
    \caption{Metrics for the model with different domain training orders}
    \label{tab:domain_ordering}
    \centering
    \begin{tabular}{c c c c}
    \toprule
        \textbf{Strategy} & \textbf{ACC} & \textbf{LA} & \textbf{FM} \\
    \midrule
        \textbf{Lowest First} & \(87.53\%\) & \(87.04\%\) & \(7.15 \times 10^{-3}\) \\ 
        \textbf{Highest First} & \(88.05\%\) & \(86.95\%\) & \(8.23 \times 10^{-3}\) \\
        \textbf{Alternated} & \(87.70\%\) & \(87.04\%\) & \(3.65 \times 10^{-3}\) \\
        \bottomrule\\
    \end{tabular}
\end{table}

\begin{figure}[!ht]
    \centering
    \includegraphics[width=0.48\textwidth]{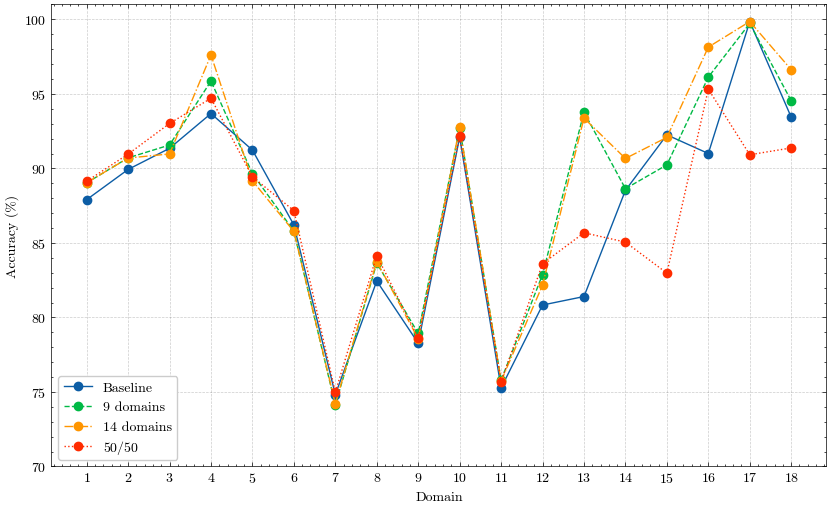}
    \caption{Accuracy with different sizes of the replay buffer}
    \label{fig:continual_exemplar}
\end{figure}

\textbf{Domain selection.} Finally, it is important to assess a ablation of the boosting-inspired experience replay strategy, focusing on how variations in replay buffer size and rehearsal policy affect forward and backward transfer in continual learning. Three different strategies were considered. Firstly, the replay buffer was restricted to a maximum capacity of 9 previously encountered domains, covering about 50\% of all domains. This setup allows us to evaluate the model's ability to retain knowledge with a moderate amount of past data. Secondly, it was restricted to a maximum capacity of 14 previously encountered domains, representing approximately 75\% of all domains. This larger buffer size provides a more comprehensive review of past tasks, potentially enhancing retention and transfer. Thirdly, a balanced selection approach was employed, where half of the replayed exemplars are drawn from domains with the lowest loss, and the other half from domains with the highest loss, alternately. Figure~\ref{fig:continual_exemplar} displays the accuracy evolution across domains, with a more detailed comparison in Table~\ref{tab:continual_exemplar} summarizing \(ACC\), \(LA\), e \(FM\) for each replay buffer size and rehearsal policy. As can be observed, increasing the size of the replay buffer improves the average accuracy (\(ACC = 88.96\%\)) but slightly increases forgetting (\(FM = 4.10 \times 10^{-3}\)). This suggests that larger buffers provide better coverage of past domains, but may introduce redundancy or noise that slightly impacts retention of earlier domain knowledge. The balanced exemplar selection strategy achieves the lowest forgetting measure (\(FM = 2.10 \times 10^{-3}\)), demonstrating its effectiveness in mitigating catastrophic forgetting by focusing equally on high-loss and low-loss domains during replay episodes. However, this strategy results in slightly lower average accuracy (\(ACC = 86.94\%\)), indicating that balancing exemplars might sacrifice some overall performance for better retention. Domain-specific trends remain consistent with previous experiments, and domain 7 shows persistent lower accuracy (\(\approx 75\%\)) regardless of the size of the replay buffer or the exemplar selection strategy, strengthening its inherent difficulty for the model to learn effectively from the characteristics of this domain.

\begin{table}[!ht]
    \caption{Metrics for the model with different sizes of the replay buffer}
    \label{tab:continual_exemplar}
    \centering
    \begin{tabular}{c c c c}
    \toprule
        \textbf{Strategy} & \textbf{ACC} & \textbf{LA} & \textbf{FM} \\
    \midrule
        \textbf{9 domains} & \(88.54\%\) & \(86.41\%\) & \(3.57 \times 10^{-3}\) \\ 
        \textbf{14 domains} & \(88.96\%\) & \(86.90\%\) & \(4.10 \times 10^{-3}\) \\
        \textbf{Balanced} & \(86.94\%\) & \(85.10\%\) & \(2.70 \times 10^{-3}\) \\
        \bottomrule\\
    \end{tabular}
\end{table}

\begin{figure}[!ht]
    \centering
    \includegraphics[width=0.485\textwidth]{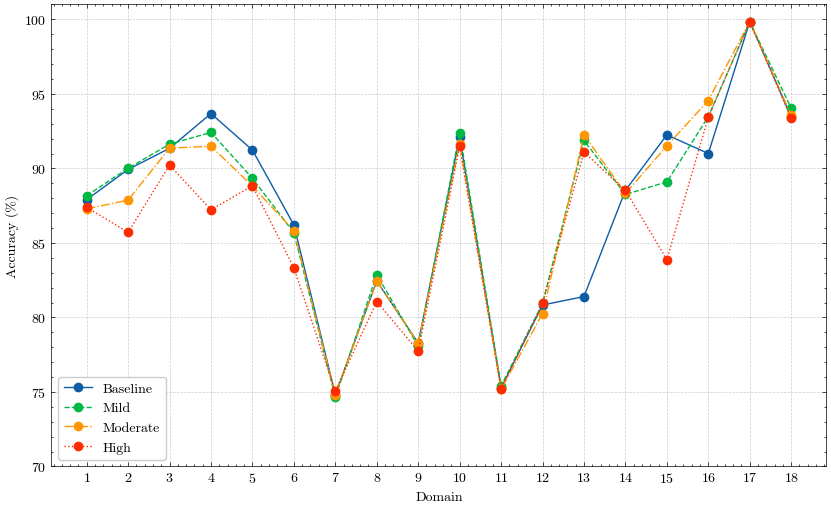}
    \caption{Accuracy with different amounts of data corruption}
    \label{fig:corruption}
\end{figure}

\textbf{Corruption robustness.} Finally, to investigate how noise influences continual learning outcomes in safety-critical scenarios, three distinct levels of corruption with adaptive data augmentation \cite{Wang2023} are evaluated, each characterized by the fraction of data chosen for corruption and the corresponding noise rate: uniform mild, selective moderate, and high-level. In uniform mild corruption, a smaller portion of the data (20\%) is uniformly selected across all domains. Each chosen instance undergoes a low noise rate (5\%). In selective moderate corruption, a moderate subset of data is used (30\%). Each chosen instance is subjected to a moderate noise rate (15\%). In high-level corruption, a larger portion of the data is targeted (40\%), and each selected instance has a higher noise rate (30\%). Figure~\ref{fig:corruption} shows the accuracy across domains for each corruption level, with a more detailed comparison in Table~\ref{tab:corrupion}. As observed, the impact of noise is evident at all levels of corruption, with mild corruption having a minimal impact on average accuracy (\(ACC = 87.67\%\)), while high corruption significantly reduces performance (\(ACC = 86.35\%\)). This highlights that the model is robust to mild noise, but struggles under severe corruption conditions. Moderate corruption achieves slightly lower average accuracy (\(ACC = 87.51\%\)), but results in better retention, as evidenced by lower FM values compared to mild corruption scenarios, suggesting that moderate noise levels may act as a form of regularization without overwhelming the model. Domain-specific observations reveal that challenging domains, such as 7 and 11, remain consistently difficult under all corruption levels, indicating that their performance issues are likely intrinsic to their data characteristics rather than being exacerbated by noise.

\begin{table}[!ht]
    \caption{Metrics for the model with different amounts of data corruption}
    \label{tab:corrupion}
    \centering
    \begin{tabular}{c c c c}
    \toprule
        \textbf{Strategy} & \textbf{ACC} & \textbf{LA} & \textbf{FM} \\
    \midrule
        \textbf{Mild} & \(87.67\%\) & \(87.28\%\) & \(4.66 \times 10^{-3}\) \\ 
        \textbf{Moderate} & \(87.51\%\) & \(87.23\%\) & \(2.89 \times 10^{-3}\) \\
        \textbf{High} & \(86.35\%\) & \(85.87\%\) & \(5.07 \times 10^{-3}\) \\
        \bottomrule\\
    \end{tabular}
\end{table}

\section{Conclusion}
\label{sec:conclusion}

The unknown dynamics of the rotating machinery generated data requires algorithms to monitor the learning process and self-diagnose changes in the context of learning. Beyond the ability to react correctly in an unfamiliar situation, these algorithms must quickly assimilate new knowledge, seeing novelty as an opportunity for learning, rather than a risk. Powerful learning can occur only if the distribution of data from the environment differs from the training data, with different cross-domain environmental and operational variations sharing the same structure both in the past and future. Thus, posing the need to address four second-order requirements beyond accuracy: catastrophic forgetting, lack of plasticity, forward transfer and backward transfer. To tackle these, the applied method involves a feature generator and overlapping domain-specific classifiers that allow for a continually growing capacity as more domains emerge, ensuring models do not interfere with each other’s plasticity, while a restricted experience replay mechanism mitigates the risk of the model forgetting, providing stability. Moreover, to leverage on the forward and backward transfer opportunities of nonlinear environmental and operational influences, domains were selectively chosen for the replay mechanism, such that each new model incorporates knowledge from the domains with the highest error in the previous episode.

\textbf{Extensive ablations.} Experiments show that the proposed continual learning method significantly enhances fault diagnosis across multi-domain environments. Specifically, the approach achieves high average domain accuracy (up to 88.96\%), competitive learning accuracy (86.90\%), and effectively mitigates catastrophic forgetting, with forgetting measures as low as \(2.70 \times 10^{-3}\). Moreover, to evaluate the robustness of the proposed method for continual learning, three sets of ablation experiments were performed. Firstly, investigating the role of domain order revealed that starting with high-performing domains establishes a strong foundation for subsequent learning (88.05\%), but slightly increases forgetting (\(8.23 \times 10^{-3}\)), while alternating between high- and low-performing domains minimizes forgetting (\(3.65 \times 10^{-3}\)), maintaining competitive accuracy (87.70\%). Secondly, a quantitative comparison of domain selection mechanisms for the boosting-inspired experience replay strategy showed that larger replay buffers improve accuracy, but may introduce redundancy, where using a bigger size resulted in an accuracy of 88.96\%, but a forgetting of \(4.10 \times 10^{-3}\). On the other hand, a balanced exemplar selection achieves good accuracy (86.94\%) with significantly lower forgetting (\(2.70 \times 10^{-3}\)). Thirdly, the model revealed to handle noise and corruption well, maintaining robust performance under mild and moderate noise levels (87.67\% and 87.51\%, respectively), while showing slight degradation under severe corruption (86.35\%), but still achieving good results.

\textbf{Research directions.} In future work, the main focus should explore how more operational and environmental variations and real-world circumstances, such as defect patterns, noise levels, speeds, and load, influence the results. These insights would be crucial for optimizing the model's ability to handle real-world CL challenges. Moreover, it is crucial to acknowledge the limitations and the necessity for validation through field trials using authentic data, uncovering unexpected outcomes, and influence of confounding factors. Naturally, the choice choosing between the classic regularization and replay-based methods that aim at capturing the common structure within various domains, and the present methodology of decomposing concepts into reusable modules with an ensemble-like representation needs to be further studied in fault diagnosis applications. Furthermore, while this work focused on such algorithm solutions, leveraging on the inductive biases of different architectural components can also yield great benefits in dealing with the stability-plasticity trade-off. This includes studying the effect of width, depth, normalization layers, skip connections and pooling layers \citep{nguyen2019architecture}, as well as training regimes, namely the effect of learning rate, batch size, dropout, activation functions, optimizer choice, dropout, weight decay, and pretraining setups \citep{mirzadeh2020understanding}.

\bibliographystyle{cas-model2-names}

\bibliography{cas-refs}


\end{document}